\documentclass[letterpaper, 10 pt, conference]{ieeeconf}  % Comment this line out if you need a4paper

\IEEEoverridecommandlockouts                              % This command is only needed if 
\usepackage{graphics} % for pdf, bitmapped graphics files
\usepackage{epsfig} % for postscript graphics files
\usepackage{mathptmx} % assumes new font selection scheme installed
\usepackage{times} % assumes new font selection scheme installed
\usepackage{amsmath} % assumes amsmath package installed
\usepackage{amssymb}  % assumes amsmath package installed
\usepackage{multirow}
\usepackage{subcaption}
\usepackage{color}
\usepackage{xspace}
\usepackage{xfrac}
\usepackage{comment}
\usepackage{url}
\usepackage{hyperref}
\usepackage{float}

\usepackage{booktabs}

\newcommand{\etal}{~et al.~}
\makeatletter
\DeclareRobustCommand\onedot{\futurelet\@let@token\@onedot}
\def\@onedot{\ifx\@let@token.\else.\null\fi\xspace}

\def\eg{\emph{e.g}\onedot} 
\def\ie{\emph{i.e}\onedot} 
 
\def\etc{\emph{etc}\onedot} 
 
\def\etal{\emph{et al}\onedot}
\makeatother

\title{\LARGE \bf
Multi-Task Domain Adaptation for Deep Learning of\\
Instance Grasping from Simulation
}

\author{Kuan Fang$^{1, 2}$, Yunfei Bai$^{1}$, Stefan Hinterstoisser$^{1}$, Silvio Savarese$^{2}$, Mrinal Kalakrishnan$^{1}$
\thanks{$^{1}$X, Mountain View, CA 94043 USA. $^{2}$Stanford University, Stanford, CA 94305 USA. This research was conducted during Kuan's internship at X.}}

\begin{document}

\maketitle
\thispagestyle{empty}
\pagestyle{empty}

%%%%%%%%%%%%%%%%%%%%%%%%%%%%%%%%%%%%%%%%%%%%%%%%%%%%%%%%%%%%%%%%%%%%%%%%%%%%%%%%
\begin{abstract}

Learning-based approaches to robotic manipulation are limited by the scalability of data collection and accessibility of labels. In this paper, we present a multi-task domain adaptation framework for instance grasping in cluttered scenes by utilizing simulated robot experiments. Our neural network takes monocular RGB images and the instance segmentation mask of a specified target object as inputs, and predicts the probability of successfully grasping the specified object for each candidate motor command. The proposed transfer learning framework trains a model for instance grasping in simulation and uses a domain-adversarial loss to transfer the trained model to real robots using indiscriminate grasping data, which is available both in simulation and the real world. We evaluate our model in real-world robot experiments, comparing it with alternative model architectures as well as an indiscriminate grasping baseline.

\end{abstract}

%%%%%%%%%%%%%%%%%%%%%%%%%%%%%%%%%%%%%%%%%%%%%%%%%%%%%%%%%%%%%%%%%%%%%%%%%%%%%%%%
\section{INTRODUCTION}
With recent progress in deep learning and reinforcement learning, large scale learning-based methods have been used in robotic manipulation~\cite{levine2016learning,gupta,kappler}. These methods enable us to replace manually designed perception and control pipelines with end-to-end neural networks learned from large training datasets. However, it is chanllenging to apply these methods on complex, real-world manipulation tasks for two reasons: First, collecting data through repeated robot experiments in the real world is time consuming. Second, evaluation and annotation of ground truth success labels can be expensive and sometimes infeasible. Although some previous works have designed experimental platforms to automatically collect and label data for non-semantic tasks such as indiscriminately grasping any object from the tray~\cite{levine2016learning}, it can quickly get more challenging in terms of data collection when switching to tasks with different settings.
\begin{figure}[ht]
\centering
\includegraphics[width=\linewidth]{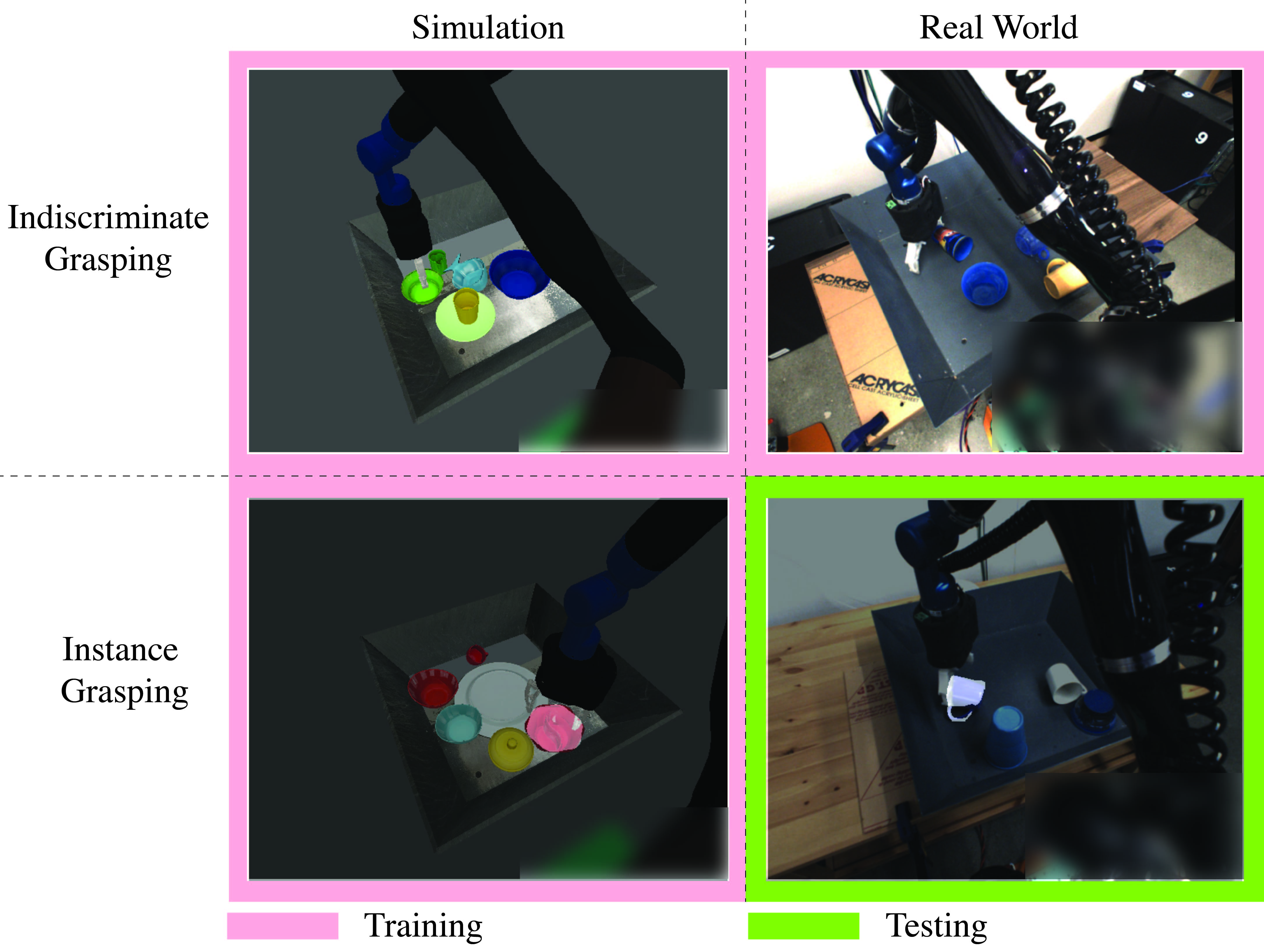}
\caption{\textbf{Multi-task domain adaptation for instance grasping.} Our multi-task domain adaptation framework trains a deep neural network for instance grasp prediction from simulation, while utilizing both real and simulated indiscriminate grasping data to mitigate the domain shift from simulation to the real world. Robot hardware details not relevant to this paper are blurred out in images to preserve confidentiality.}
\label{fig:intro}
\vspace{-6mm}
\end{figure}

One alternative solution to large scale data collection in robotics is to run robot experiments in simulation~\cite{saxena2008robotic,rusu2016sim,tobin2017domain,james2017transferring,mahler2017dex,viereck2017learning}. Simulated robot experiments are parallelizable and can easily be reset. In addition, simulation also provides access to the ground-truth states of the robot and environment, making it straightforward to generate labeled data. However, learning from only simulated data does not yield robust real-world performance due to the reality gap between the real world and the simulation in terms of robotic perception and physics. In general, due to the phenomenon known as \textit{domain shift}~\cite{quinonero2008covariate}, a model trained on data from one domain does not generalize well to another target domain. A common approach to mitigate the domain shift is to use domain adaptation~\cite{pan2010survey} to transfer the learned knowledge from one domain to another. In our context of transferring the learned policy from simulation to the real world (sim-to-real transfer), this transfer can be achieved by feeding both real and simulated data into the neural network and applying a similarity loss to regularize the difference between extracted features from the two domains. During this process, one usually still needs to collect sufficient amount of real-world data in order to learn the target task.

In this paper, we aim to address the problem of sim-to-real transfer where there is no labeled data available in the real world. Specifically, we address this problem in the context of \textit{instance grasping}, where the goal is to grasp a particular object instance from a cluttered scene. It is more challenging in terms of both perception and control compared to what we refer to as \textit{indiscriminate grasping}, where the robot can grasp any objects from the workspace indiscriminately. More importantly, unlike recent works on indiscriminate grasping~\cite{levine2016learning,gupta}, to get labeled real-world data for instance grasping, we would need sophisticated perception system to robustly identify and track the target object across time. To overcome this hurdle, we propose to only use simulation to collect labeled data for instance grasping. Furthermore, to resolve the domain shift issue, we propose a multi-task domain adaptation framework, which trains a deep neural network for instance grasp prediction from simulation, while utilizing both real and simulated indiscriminate grasping data to mitigate the domain shift (Fig.~\ref{fig:intro}). The intuition of using indiscriminate grasping data is that indiscriminate and instance grasping share similar properties regarding both robot action space and perception. This framework shares model parameters between indiscriminate grasp prediction and instance grasp prediction, and trains both tasks simultaneously. Meanwhile, it enforces the extracted features to be transferable between the simulation and the real world through domain adaptation.

Our method is demonstrated on robot instance grasping using RGB images as inputs, which are captured from a moving monocular RGB camera mounted on the robot arm. We use Mask R-CNN~\cite{he2017mask} to get segmentation masks for object instances, one of which is then sampled to specify the target object for grasping. Our method only relies on a single segmentation mask obtained at the beginning of the grasp, rather than requiring the segmentation mask at each time-step. We evaluate our method by showing the robot grasping a variety of household dishware objects, as well as its generalization to unseen objects during instance grasp training in simulation.

Our main contributions are:
\begin{itemize}
    \item A multi-task domain adaptation framework that trains a model for instance grasping in simulation and uses a domain-adversarial loss to transfer the trained model to real robots using indiscriminate grasping data, which is available both in simulation and the real world.
    \item A deep neural network architecture which takes monocular RGB images and the instance segmentation mask of a specified target object as inputs, and predicts the probability of successfully grasping the specified object for each candidate motor command.
    \item We demonstrate our method on a robot platform with a mounted moving monocular RGB camera. We showed instance grasping with a variety of household dishware objects, and the generalization to unseen objects of instance grasp training in simulation.
\end{itemize}

The rest of this paper is laid out as follows. In Section II we discuss related work. In Section III, we introduce the necessary background in end-to-end grasp prediction and domain adaptation upon which our work is based. We describe our main algorithmic contributions in Section IV. Experimental method and results on real robots are presented in Section V. Finally, we discuss potential directions for future work in Section VI.

%%%%%%%%%%%%%%%%%%%%%%%%%%%%%%%%%%%%%%%%%%%%%%%%%%%%%%%%%%%%%%%%%%%%%%%%%%%%%%%%
\section{RELATED WORK}
\begin{figure*}[ht]
\centering
\includegraphics[width=\linewidth]{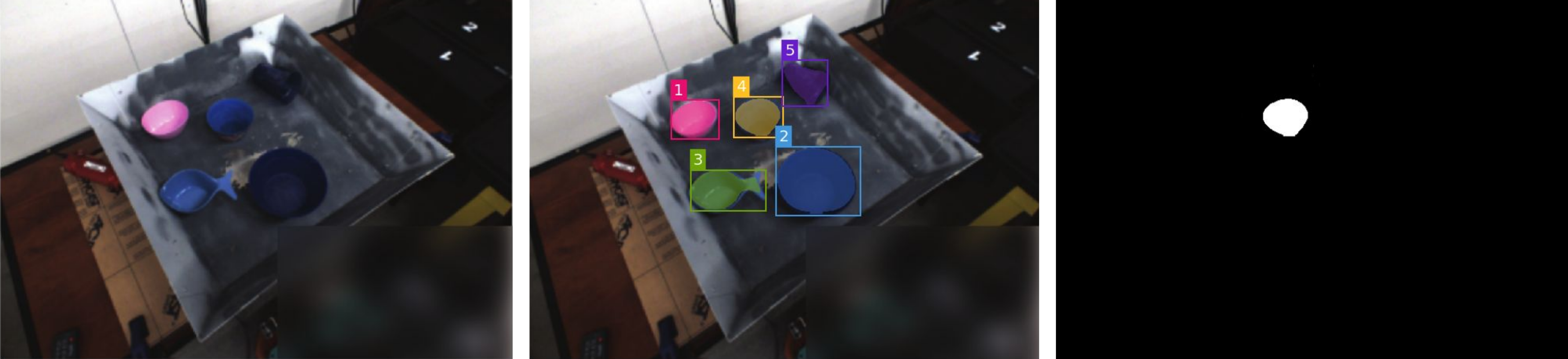}
\caption{\textbf{The pipeline of obtaining the segmentation mask for a target object.} Given the initial camera image (left), we forward the Mask R-CNN to predict the instance segmentation masks (middle). The detection bounding boxes and segmentation masks of different object instances are shown in different colors, with the assigned object IDs shown on the upper left corner. Then one target object is sampled and its segmentation mask is defined as the target mask (right).}
\label{fig:mask_rcnn_pipeline}
\vspace{-6mm}
\end{figure*}

\textbf{Robotic grasping.} Grasping is one of the most fundamental robot manipulation tasks. Both geometry-based~\cite{force_closure_1995,caging_servey,bai2014dexterous} and data-driven methods~\cite{bohg2014data} have been proposed to address the problem of robotic grasping. Recent works~\cite{levine2016learning,gupta} have used deep learning to train models that can predict grasp success directly from visual sensor inputs, which has the benefit of generalization. Other works focused on semantic grasping~\cite{jang2017end}, where the robot picks up an object of a specified class. In our work, we propose a deep learning algorithm for instance grasping, and we use simulation to generate labeled instance grasping data.

\textbf{Transfer learning from simulation to real robots.} Deep learning methods for robotic manipulation usually require large amounts of data on the order of a million trials~\cite{levine2016learning}. Recent works have considered using simulation to train robotic manipulation to scale up the data collection. Saxena \etal~\cite{saxena2008robotic} used rendered objects to learn a vision-based grasping model. Rusu \etal ~\cite{rusu2016sim} introduced progressive neural networks to adapt an existing deep reinforcement learning policy trained for a reaching task in simulation to the real world. However, due to differences in the dynamics and observation models (reality gap), solving the problem of sim-to-real transfer is the key to ensuring that policies trained in simulation can work in the real world~\cite{taylor2009transfer}.

One way for addressing sim-to-real transfer is to choose input modalities where the reality gap is not significant. One option is to use the depth image which abstracts away many appearance properties of real-world objects that are hard to render. A number of recent works used simulated depth images to learn indiscriminate grasping with deep convolutional neural networks (CNNs)~\cite{mahler2017dex,viereck2017learning}, then deploy the trained policy on the real robot with a calibrated fixed depth camera or a wrist mounted depth camera for indiscriminate grasping. Our work tackles instance grasping problem which is much more difficult. In addition, we use only monocular RGB images, without relying on either camera calibration or object geometry information. We believe there is considerable value in studying grasping systems which use RGB images only, considering the relatively low cost of monocular RGB cameras and the limitations of depth cameras (\eg, unsuitable for bright outdoor lighting, transparent object, \etc).

Some recent works have shown the success of using domain randomization for transferring a deep neural network trained on simulated RGB images to the real world for solving end-to-end robot manipulation or mobility tasks~\cite{sadeghi2016cad2rl,tobin2017domain,james2017transferring}. These works extended prior works on data augmentation in computer vision~\cite{simard2003best}, and applied randomization to parameters such as camera position, lighting, and texture in simulation. However, only simple geometries like cubes or free space motions are considered in those prior methods. In contrast, we address the problem of grasping diverse naturalistic household objects like bowls and cups, which are much more complex in terms of both perception and contact physics. Furthermore, we also demonstrate sim-to-real transfer with generalization to new objects that were never seen during training in simulation.

Domain adaptation is a process that allows a learning model trained with samples from a source domain, in our case the simulation domain, to generalize to a target domain, the real world. Feature-level domain adaptation focuses on learning domain-invariant features, either by learning a transformation of fixed, pre-computed features between source and target domains~\cite{sun2015return,gong2012geodesic,caseiro2015beyond,gopalan2011domain} or by learning a domain-invariant feature extractor, often as a CNN~\cite{ganin2016domain, long2015learning, tzeng2015ddc, bousmalis2016domain}. Tzeng \etal~\cite{tzeng2016adapting} perform sim-to-real transfer for a manipulation task using a adversarially trained domain discriminator, by doing nearest neighbor in the learned feature space to find weak pairings. A recent study~\cite{bousmalis-2018} used combined feature-level and pixel-level domain adaption for learning robotic hand-eye coordination for grasping a variety of everyday objects, by training primarily with simulation data and small amount of labeled real-world data. Our work uses feature-level domain adaptation, particularly we propose a multi-task domain adaption method to address the lack of labeled data for instance grasping in the real world.
%To our knowledge, this is the first time... (no real-world label, instance grasping, dishware objects, generalization, multi-task domain adaptation) 

\textbf{Instance segmentation and object detection.}
Object detection and segmentation have been a long standing computer vision problem. Several recent deep learning techniques \cite{he2017mask, Girshick2014RichFH, He2014SpatialPP, Girshick2015FastR, ren2015faster, Redmon2016YouOL, Shelhamer2015FullyCN, Chen2017DeepLabSI} have made significant progress in this direction. In our work, we use Mask R-CNN \cite{he2017mask} trained on synthetic images to segment objects from the background in real-world images. This gives us an easy way to specify the object to grasp without manually labeling objects in the image. Our proposed method could potentially also be trained to use a bounding box or a few pixels on the object as input instead of the instance mask. However, exploration of these alternative input modalities is left for future works.

%%%%%%%%%%%%%%%%%%%%%%%%%%%%%%%%%%%%%%%%%%%%%%%%%%%%%%%%%%%%%%%%%%%%%%%%%%%%%%%%
\section{BACKGROUND}
\begin{figure*}[ht]
\centering
\includegraphics[width=\linewidth]{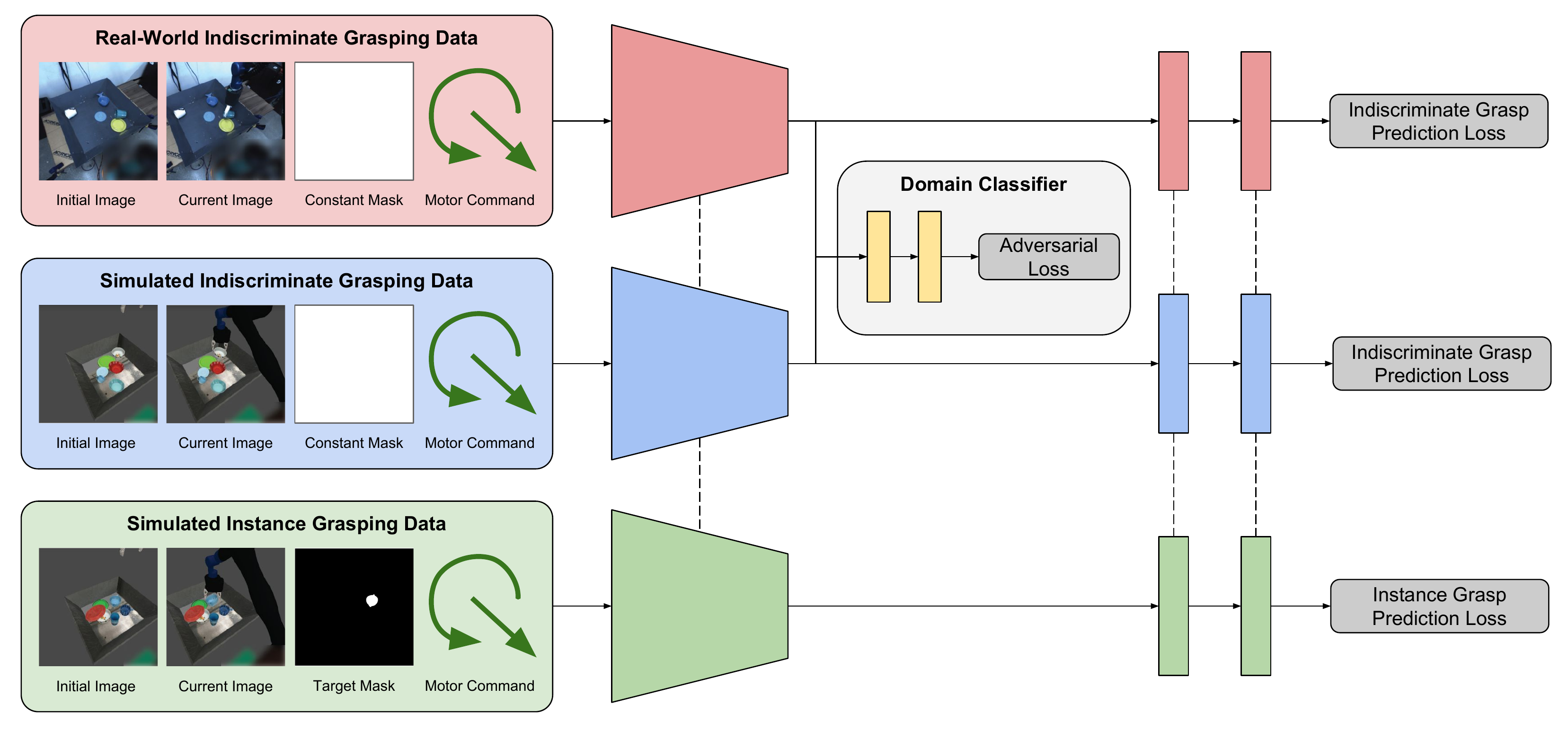}
\caption{\textbf{Multi-task domain adaptation framework.} The framework is composed of three grasp prediction towers and a domain classifier. Each tower takes training data from one of the three task domains: real-world indiscriminate grasping, simulated indiscriminate grasping and simulated instance grasping. The neural networks for the three towers share parameters as denoted by the dashed lines. In this framework, the neural network is trained to predict the instance grasp success probability with transferable features between the simulation and the real world by minimizing the four losses simultaneously with respect to the network parameters.}
\vspace{-4mm}
\label{fig:domain_adaptation_framework}
\end{figure*}
%%%%%%%%%%%%%%%%%%%%%%%%%%%%%%%%%%%%%%%%%%%%%%%%%%%%%%%%%%%%%%%%%%%%%%%%%%%%%%%%
\subsection{End-to-End Grasp Prediction}
In this work, we follow the formulation and real-world experimental setup of end-to-end grasp prediction described in \cite{levine2016learning}. The goal is to choose a sequence of robot actions step by step given the current input RGB image, in order to control the robot gripper to grasp any object from the cluttered scene. We denote the grasp prediction network as $g(I_0, I_t, v_t; \theta)$, where $I_0$ and $I_t$ are the input RGB images from the robot camera at the initial time step and the current time step, $v_t$ is a given robot command sampled from the action space, $\theta$ represents the neural network parameters. At each time step during grasping, the network predicts the probability of grasp success for a set of sampled actions, and chooses the action with the highest predicted probability of success. We run the cross-entropy method (CEM)~\cite{rubinstein2004cross} for 3 iterations to find action sampling distributions centered around regions of high grasp success probability. In each iteration, we fit a single Gaussian distribution to the top-ranked motor commands. A grasp is executed when the predicted grasp success probability is above a threshold.

The dataset for training the grasp prediction network is collected by controlling the gripper to repeatedly attempt to grasp objects from a tray with randomly selected objects and drop them back. This process starts with exploring a random policy at the beginning, and then switches to the CEM with the updated network parameters. In the real world, the ground truth success label is automatically determined by a hard coded perception system after each grasping episode. This is done by taking images of the tray and comparing pixel differences before and after dropping the object. In simulation, we determine the success label by checking the object position. The grasp prediction network is then trained using a log loss against the ground truth success labels.

%%%%%%%%%%%%%%%%%%%%%%%%%%%%%%%%%%%%%%%%%%%%%%%%%%%%%%%%%%%%%%%%%%%%%%%%%%%%%%%%
\subsection{Adversarial Loss for Domain Adaptation}
We use the adversarial loss \cite{ganin2016domain} during training to penalize the domain shift from simulation to the real world similar as in the recent study~\cite{bousmalis-2018}. The adversarial loss regularizes the neural network weights $\theta$ through a domain classifier $D(I_0, I_t, v_t; \theta, \phi)$, where $\phi$ represents the parameters of the domain classifier. In our problem setup, the target domain is the real world and the source domain is simulation. During training, data from the two different domains are fed into the grasp prediction network. The domain classifier takes as input the intermediate feature extracted from the neural network, and is trained to predict which domain the feature is from, by maximizing the binomial cross-entropy~\cite{ganin2016domain} with respect to $\phi$. Meanwhile we minimize the adversarial loss with respect to $\theta$: 
\begin{equation}
    \mathcal{L}_{adversarial} = \sum_{i = 0}^{N_S + N_R} d^i \log \hat{d^i} + (1 - d^i) \log (1 - \hat{d^i})
\end{equation}
where $d^i \in \{0, 1\}$ is the ground truth domain label for each input $I_0^i$, $I_t^i$, $v_t^i$ of sample $i$, $\hat{d^i} = D(I_0^i, I_t^i, v_t^i; \theta, \phi)$ is the predicted domain label, $N_S$ and $N_R$ are the sizes of the simulated and real datasets.

%%%%%%%%%%%%%%%%%%%%%%%%%%%%%%%%%%%%%%%%%%%%%%%%%%%%%%%%%%%%%%%%%%%%%%%%%%%%%%%%
\section{Multi-Task Transfer Learning from Simulation for Instance Grasping}

%%%%%%%%%%%%%%%%%%%%%%%%%%%%%%%%%%%%%%%%%%%%%%%%%%%%%%%%%%%%%%%%%%%%%%%%%%%%%%%%
\subsection{End-to-End Instance Grasp Prediction}
\label{sec:instance_grasp_prediction}
The goal of instance grasp prediction is to predict the probability of successfully grasping the specified object for each candidate motor command. The target object is specified by a segmentation mask $M_0$ given at the beginning of the episode before grasping starts. To predict the success probability for instance grasping, we define the instance grasp prediction network $g(M_0, I_0, I_t, v_t; \theta)$ by extending the formulation for indiscriminate grasping~\cite{levine2016learning}. In addition to RGB images and robot commands, the instance grasp prediction network also uses the segmentation mask $M_0$ of the target object as input. In order to predict the instance grasp success given each sampled action, the network is required to understand the relative location between the gripper and the target object by fusing this information together.

For training, we only collect instance grasping data in simulation. It is crucial to keep the labels of successful and failed trials balanced during training. Directly running a random policy for data collection yields very few successful trials. To improve sample efficiency, we use a hindsight data collection trick inspired by \cite{Andrychowicz2017HindsightER}. Here we first run indiscriminate grasping data collection as described in~\cite{levine2016learning}. For each successful indiscriminate grasping trial, we generate one successful instance grasping trial by taking the mask of the grasped object as the input target mask, and one failed instance grasping trial by using a mask sampled from other objects. As a result, about 20\% of the grasping trials in our training dataset are labeled as success for the task of instance grasping. During test time, instance grasping results are manually evaluated by human annotators.

Our pipeline to get a sampled segmentation mask is shown in Fig.~\ref{fig:mask_rcnn_pipeline}. Given an initial RGB image $I_0$ at the test time, we first run Mask R-CNN\cite{he2017mask} to predict the instance segmentation masks, where each mask is assigned to an unique object instance ID. Then we sample one of the object IDs and get the corresponding segmentation mask $M_0$ for the target object. Note that we only rely on the \textit{initial} segmentation mask as the input to the instance grasp prediction neural network. Since the robot camera is attached to a pan-tilt unit mounted on the first link of the arm, it results viewpoint changing as the arm moves around. Therefore, $M_0$ does not align with the target object in the following RGB images $I_t$. However, we still find $M_0$ is effective to provide information about the location of the target object, by fusing both the segmentation and RGB images as input to the neural network, as discussed in Section~\ref{architecture}.

%%%%%%%%%%%%%%%%%%%%%%%%%%%%%%%%%%%%%%%%%%%%%%%%%%%%%%%%%%%%%%%%%%%%%%%%%%%%%%%%
\subsection{Multi-Task Domain Adaptation}
\begin{figure}[ht]
\centering
\includegraphics[width=\linewidth]{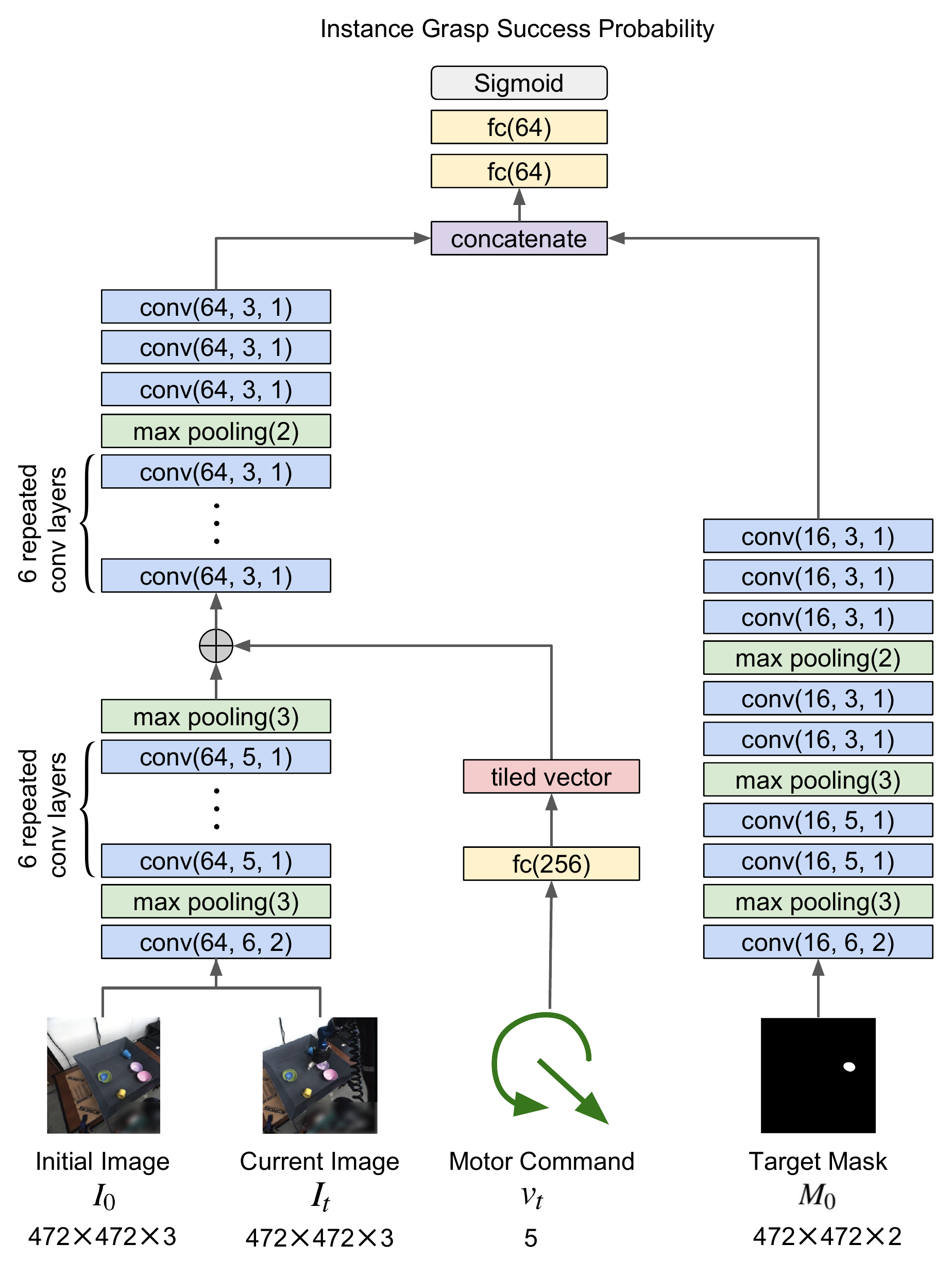}
\caption{\textbf{Instance Grasp Prediction Network.} Taking the initial and current camera images, the candidate motor command and the target mask as inputs, the network predicts the instance grasp success probability for the candidate motor command. Convolutional layer (conv) parameters are shown as (number of filters, kernel size, stride). Fully-connected layer (fc) parameters are shown as (number of channels). Max pooling layer parameters are shown as (stride).} 
\label{fig:neural_network}
\vspace{-6mm}
\end{figure}
Our multi-task domain adaptation framework is shown in Fig.~\ref{fig:domain_adaptation_framework}. The framework is composed of a domain classifier and three grasp prediction towers with training data from three task domains: real-world indiscriminate grasping data $\mathbf{X}_{R}^{A}$, simulated indiscriminate grasping data $\mathbf{X}_{S}^{A}$, and simulated instance grasping data $\mathbf{X}_{S}^{B}$. Here for simplicity of notation, we use $S$ and $R$ to represent the simulation and the real world, and use $A$ and $B$ to represent indiscriminate grasping and instance grasping respectively. The three grasp prediction towers are trained simultaneously by minimizing the following loss with respect to the neural network parameters $\theta$:
\begin{equation}
    \mathcal{L} = \mathcal{L}_{S}^{B} + \alpha \mathcal{L}_{R}^{A} + \beta \mathcal{L}_{S}^{A}  + \lambda \mathcal{L}_{adversarial}
\end{equation}
\label{eqn:total_loss}
where $\lambda$, $\alpha$ and $\beta$ are weights of the loss terms. The instance grasp prediction loss $\mathcal{L}_{S}^{B}$ trains the neural network to predict instance grasp success probability in the simulation. The two indiscriminate grasp prediction loss $\mathcal{L}_{R}^{A}$ and $\mathcal{L}_{S}^{A}$ train the neural network to extract meaningful features for grasp prediction both in the simulation and the real world. The adversarial loss $\mathcal{L}_{adversarial}$ regularizes the neural network parameters through the domain classifier. By attempting to confuse the domain classifier, the neural network learns to extract features that are transferable between the simulation and the real world.

The three towers use the same instance grasp prediction network architecture and share all network parameters. Since the segmentation mask input is only available in the instance grasping task, we feed in a constant mask to the two indiscriminate grasping towers. The constant mask marks every pixel as a candidate object to be grasped. The interpretation of the mask is that any object covered by the mask can be grasped by the robot in order to achieve success, and this interpretation is consistent across both the instance grasping and indiscriminate grasping tasks. This enables us train on the three task domains with the same input modalities and avoid training additional layers only for the simulated instance grasping data.

%%%%%%%%%%%%%%%%%%%%%%%%%%%%%%%%%%%%%%%%%%%%%%%%%%%%%%%%%%%%%%%%%%%%%%%%%%%%%%%%
\subsection{Neural Network Architecture} \label{architecture}
\begin{figure}[t]
\centering
\includegraphics[width=1.0\linewidth]{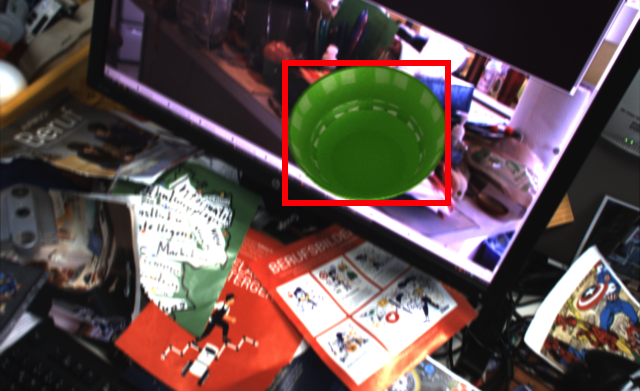}
\caption{\textbf{Representative synthetic image for training the Mask R-CNN.} Each image consist of a real background image and the objects rendered on top of it. The ground truth detection bounding box and the segmentation mask are overlaid on the rendered object.}
\label{fig:mask_training_image}
\vspace{-6mm}
\end{figure}
We show the architecture details of our instance grasp prediction network $g(M_0, I_0, I_t, v_t; \theta)$ in Fig.~\ref{fig:neural_network}. The original RGB images and target masks are $640 \times 512$ and we cropped them into $472 \times 472$ as input to the network. The candidate gripper motor command vector has 5 dimensions: a 3D translation vector, and a sine-cosine encoding of the change in orientation. We adopt the same layers from \cite{levine2016learning} to extract low level features of RGB images and motor commands. To extract the target mask features, we introduce a separate stream of convolutional layers. This mask stream maps the target mask to a convolutional feature which matches the spatial size of the features from the RGB stream. The features from the two streams are concatenated at the final convolutional layer and then two fully connected layers merge the information to predict instance grasp success probability.

To have fast convergence rate during training, convolutional layers and fully connected layers usually need batch normalizations~\cite{ioffe2015batch}. However, as discussed in \cite{bousmalis-2018}, sharing batch normalization parameters across domains can significantly harm the performance during testing in a domain adaptation framework. This is due to the inconsistent behaviors of batch normalization layers during training and testing. To resolve this problem, we remove all batch normalization layers in the grasp prediction network. Instead we use instance normalization~\cite{ulyanov2016instance} for convolutional layers and layer normalization~\cite{Ba2016LayerN} for fully connected layers, which also guarantees fast convergence rate but have consistent behaviors between training and testing.

%%%%%%%%%%%%%%%%%%%%%%%%%%%%%%%%%%%%%%%%%%%%%%%%%%%%%%%%%%%%%%%%%%%%%%%%%%%%%%%%
\section{EXPERIMENTS}
%%%%%%%%%%%%%%%%%%%%%%%%%%%%%%%%%%%%%%%%%%%%%%%%%%%%%%%%%%%%%%%%%%%%%%%%%%%%%%%%
\subsection{Instance Segmentation Prediction with Mask R-CNN}
In order to provide instance detection and segmentation, we apply Mask R-CNN~\cite{he2017mask} to the initial input image.
Mask R-CNN is trained only with synthetically generated images~\cite{hinterstoisser2017pre}, \ie images which
consist of a real background image and the objects rendered on top of it (shown in Fig.~\ref{fig:mask_training_image}). We ensure that the training set uniformly
covers the whole pose space of interest, including all possible translations and rotations under which we could potentially observe the objects.
Although occlusion is not explicitly covered in the training set, the trained Mask-RCNN demonstrates the capability of detecting and segmenting object instances from cluttered scenes.

%%%%%%%%%%%%%%%%%%%%%%%%%%%%%%%%%%%%%%%%%%%%%%%%%%%%%%%%%%%%%%%%%%%%%%%%%%%%%%%%
\subsection{Data Collection and Training}
\begin{figure}[t]
\centering
\includegraphics[width=\linewidth]{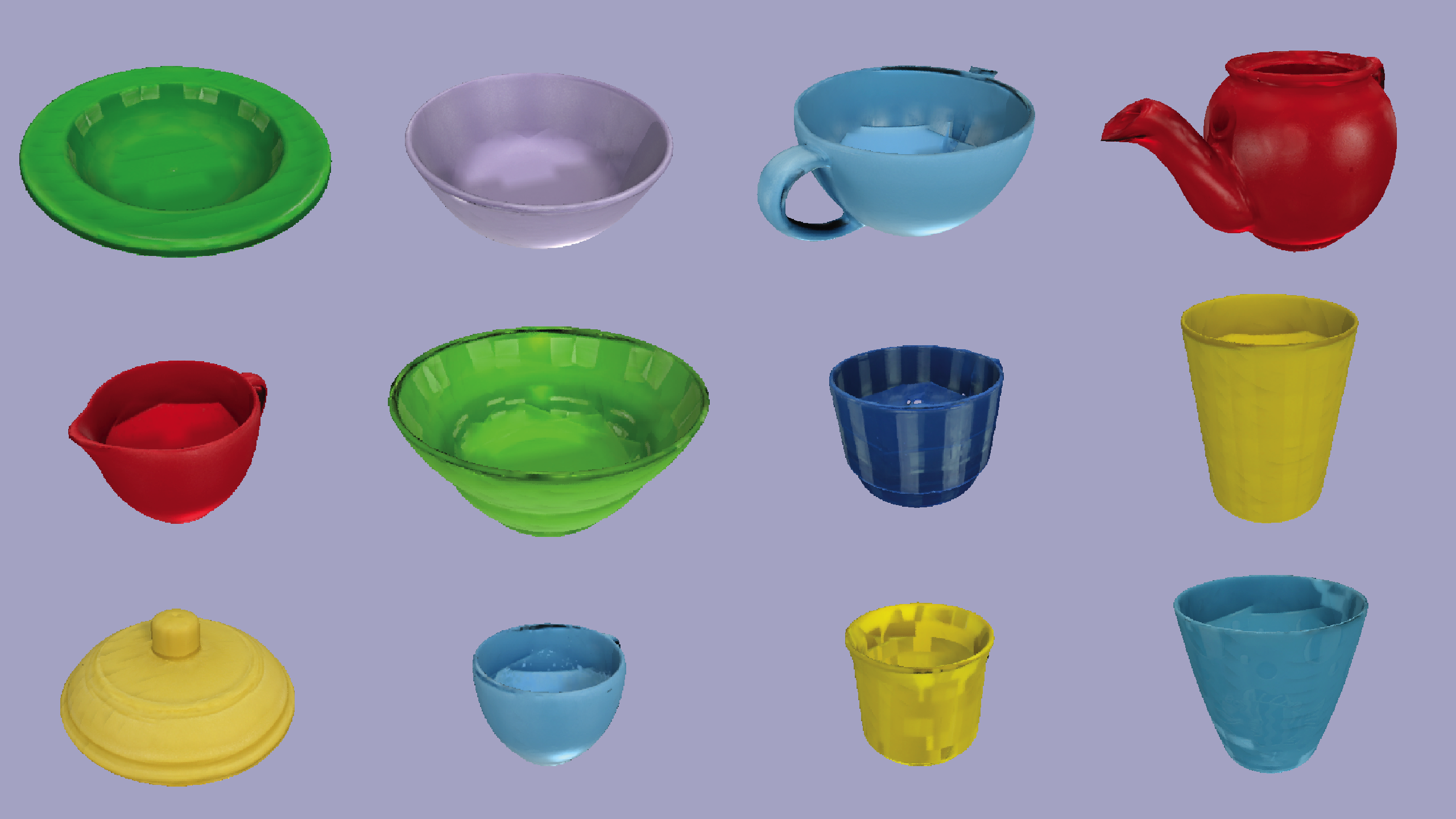}
\caption{\textbf{Examples of our 3D models used for data collection in simulation.} We use totally 130 3D models scanned from real-world dishware objects including bowls, mugs, cups, teapots \etc. The object textures are rendered in simulation.}
\label{fig:training_object}
\vspace{-4mm}
\end{figure}
To collect real-world indiscriminate grasping data, we use an automatic data collection platform similar to \cite{levine2016learning}. We ran indiscriminate grasping on 9 Jaco robot arms, with a random policy as well as a CEM policy using a trained model, to collect 100,000 grasping trials on a total of around 100 unique dishware objects. We use standard data augmentation methods including random cropping and random image distortion to increase data diversity.

To collect both instance and indiscriminate grasping data in simulation, a basic virtual environment is built based on the Bullet physics simulator~\cite{coumans2013bullet} and a simple software renderer shipped with Bullet is used for rendering. We scanned 130 3D models of household dishware objects (\eg mugs, bowls, water bottles) as the objects to be grasped in simulation (shown in Fig.~\ref{fig:training_object}). The environment emulates the same Jaco robot arm setup by simulating the physics of grasping and by rendering what the moving camera mounted on the robot would perceive: the robot arm and gripper, the bin that contains the objects, and the dishware objects to be grasped.
We collect 1 million indiscriminate grasping trials using a random policy and the CEM policy on iteratively trained models, with 1 to 6 objects in the tray per episode.
The data is then labeled post-hoc for instance grasping as described in Sec.~\ref{sec:instance_grasp_prediction}.

We train on 10 GPUs for 300k iterations with a learning rate of $0.0001$ which is decayed by 0.94 every 50k iterations, and a momentum of 0.9. We use a batch size of 32 for data from each of the three task domains. The training starts with only the three grasp prediction losses and the adversarial loss is added after 50k iterations The weights of the loss terms in \ref{eqn:total_loss} are chosen as $\alpha = 1$, $\beta = 1$, $\gamma = 4$. All of the hyperparameters are selected through experiments in \cite{bousmalis-2018} without tuning on any instance grasping validation datasets.

%%%%%%%%%%%%%%%%%%%%%%%%%%%%%%%%%%%%%%%%%%%%%%%%%%%%%%%%%%%%%%%%%%%%%%%%%%%%%%%%
\subsection{Evaluation of Instance Grasping}
\label{sec:evaluation}
\begin{figure}[t]
\centering
\includegraphics[width=.55\linewidth]{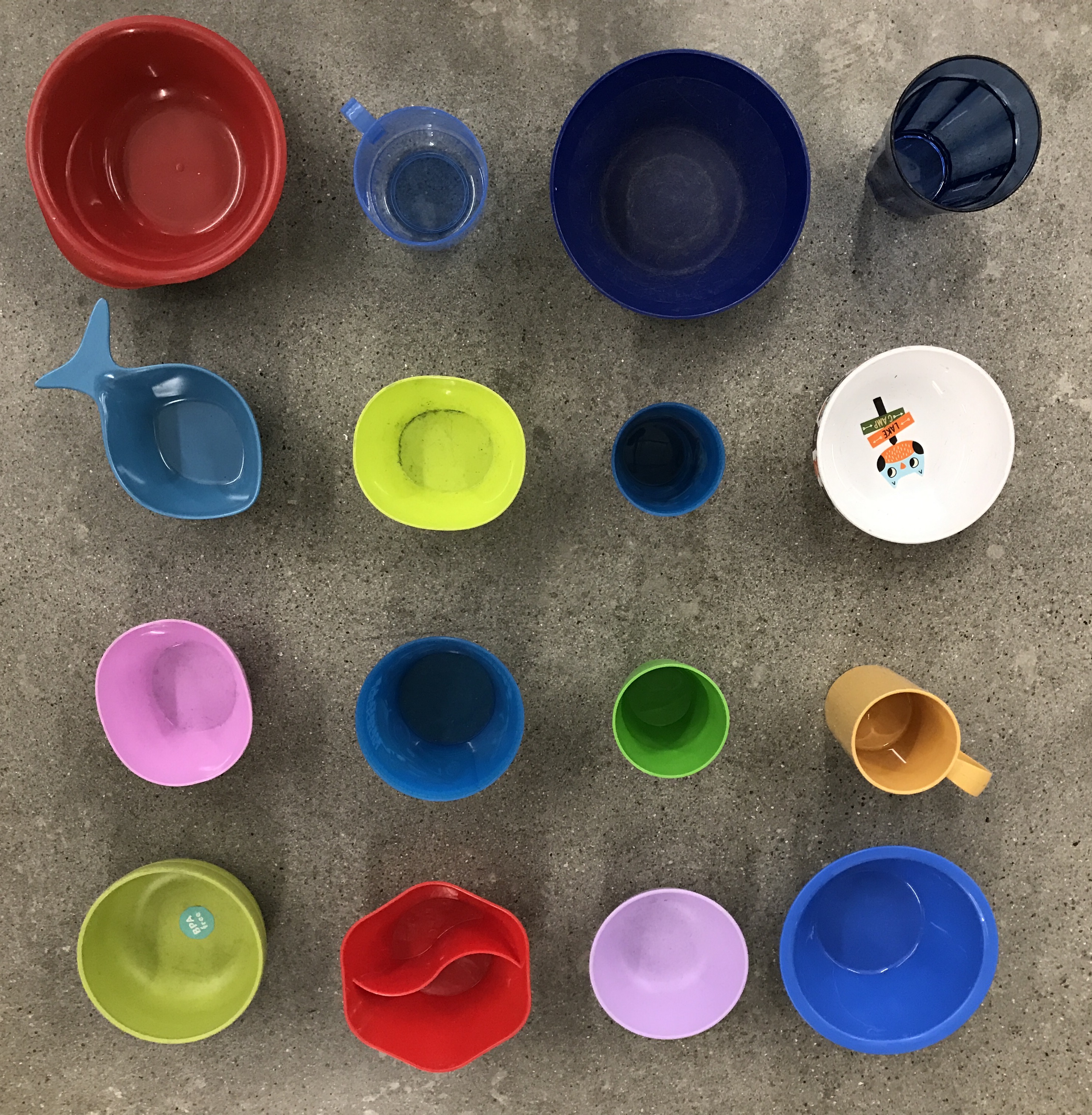}
\caption{\textbf{Our testing objects in the real world.} In each grasping trial, a combination of 5 objects is chosen and put into the tray in front of the robot. The target object is sampled from the detected objects in the tray using the Mask-RCNN.}
\label{fig:testing_objects}
\vspace{-4mm}
\end{figure}
To evaluate the instance grasping performance in the real world, we run 250 trials for each method across 5 Jaco robot arms. In each trial we place a combination of 5 objects in the tray. We use the same 10 object combinations chosen from 16 dishware objects with various colors and shapes as shown in Fig.~\ref{fig:testing_objects}. The objects chosen are present in the real-world \textit{indiscriminate} grasping dataset, however they are held out from the \textit{instance} grasping dataset collected in simulation. Table~\ref{tab:grasp_success} shows the detail of the results. Our framework achieves 60.8\% success rate for instance grasping. Among the failed instance grasping attempts, 26 attempts ended up with grasping the wrong object, 72 attempts failed grasping anything from the tray. Further
details and videos can be found at \href{https://sites.google.com/view/multi-task-domain-adaptation/}{https://sites.google.com/view/multi-task-domain-adaptation/}

%%%%%%%%%%%%%%%%%%%%%%%%%%%%%%%%%%%%%%%%%%%%%%%%%%%%%%%%%%%%%%%%%%%%%%%%%%%%%%%%
\begin{table*}[!t]
\centering
\caption{\textbf{Performance of instance grasping in real-world experiments.} For each method, we run 250 trials across 5 Jaco robot arms. We show numbers of successful instance grasps, wrong object grasps, failed grasps, and the instance grasping success rate. Our method achieves the highest instance grasping success rate among the four as described in \ref{sec:evaluation} and \ref{sec:ablation}.}
\begin{tabular}{ l | c | c | c | c }
\hline
{\bf Method} & \bf Successful Instance Grasps &\bf Grasped Wrong Objects & \bf Failed Grasps & \bf Instance Grasping Success Rate \\ 
\hline
Indiscriminate & 34 & 169 & \textbf{47} & 13.6\%\\ \hline
% Frozen Weights  &  &  &  & \%\\ \hline
Two-Tower  & 87 & 31 & 132 & 34.8\%\\ \hline
No Constant Mask & 146 & \textbf{24} & 80 & 58.4\%\\ \hline
Our Method & \textbf{152} & 26 & 72 & \textbf{60.8\%}\\ \hline
\end{tabular}
\label{tab:grasp_success}
\vspace{-4mm}
\end{table*}
\begin{figure}[ht]
\centering
\includegraphics[width=1.0\linewidth]{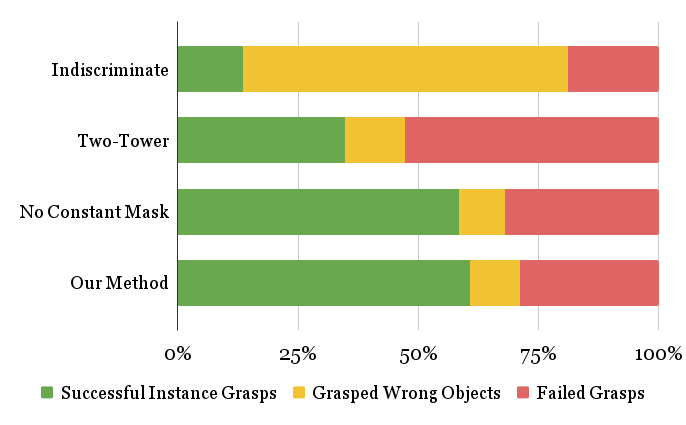}
\caption{\textbf{Failure Analysis.} We show the breakdown of successful instance grasps, wrong object grasps, and failed grasps for each of the methods described in \ref{sec:evaluation} and \ref{sec:ablation}.}
\label{fig:failure_analysis}
\vspace{-4mm}
\end{figure}
%%%%%%%%%%%%%%%%%%%%%%%%%%%%%%%%%%%%%%%%%%%%%%%%%%%%%%%%%%%%%%%%%%%%%%%%%%%%%%%%
\subsection{Ablative Analysis}
\label{sec:ablation}
We run a number of ablations with the same evaluation settings to analyze the instance grasping results of multi-task domain adaptation framework. Detailed comparisons are shown in Table~\ref{tab:grasp_success} and Fig.~\ref{fig:failure_analysis}.

\textbf{Indiscriminate vs.\ Instance Grasp Prediction Network:} The instance grasp prediction network controls the gripper to grasp the specified target object. While the indiscriminate grasp prediction network does not distinguish objects. To show this difference, we run indiscriminate grasping trials using a network trained only for indiscriminate grasp prediction with the sim-to-real domain adaptation algorithm for single task~\cite{bousmalis-2018}. With this indiscriminate policy the robot gripper successfully grasps objects from the tray in 81.2\% of the trials. Since no targets are specified, many grasps only aim for easy objects. To show the discrepancy of the objectives between the two tasks, we also sample target masks and evaluate if the grasp hits the target in each trial as in the instance grasping task. Only 13.6\% of the trials end up grasping the correct target objects.

\textbf{Two-Tower vs.\ Three-Tower:} The transferring ability of the adversarial loss comes from training the network to confuse the adversarial loss. The domain shift can be difficult to resolve when using only the real world indiscriminate grasping data and simulated instance grasping data, since both the objective and the perception for the two task domains are different. To demonstrate this, we take the two-tower domain adaptation framework from \cite{ganin2016domain} using data only from these two task domains. The adversarial loss is applied to the real world indiscriminate grasping data and simulated instance grasping data, without using simulated indiscriminate grasping data to bridge the gap. The instance grasp success rate of the two-tower framework is 34.8\%. Comparing with the three-tower model, the two-tower model failed in grasping any objects in more trials. This suggests that our framework transfers the learned instance grasping policy better than the two-tower framework.  

\textbf{Using Constant Masks vs.\ Training Additional Layers:} 
When dealing with the different input modalities between indiscriminate and instance grasping, one alternative to using constant masks is using different network architectures for the two tasks. However, additional layers have to be used for instance grasping since the extracted features can be very different with and without the information of the target mask. We thus modify our framework by only sharing the convolutional layer parameters of the RGB stream between the instance and indiscriminate networks, but train new fully connected layers for instance grasping. With layers only trained in simulation, this framework achieves 58.4\% grasping success rate for instance grasping. This shows the performance gains of introducing constant masks for using same network architectures.

%%%%%%%%%%%%%%%%%%%%%%%%%%%%%%%%%%%%%%%%%%%%%%%%%%%%%%%%%%%%%%%%%%%%%%%%%%%%%%%%
\section{CONCLUSION}

%\yunfei{We can talk about how our multi-task domain adaptation can be generalized to other tasks.}
%\kalakris{Also talk about bounding box inputs or pixel inputs maybe? Because I mentioned this somewhere else in the paper and said that we do not consider these input modalities here, and it is left for future work.}

In this work, we introduce a multi-task domain adaptation framework for instance grasping in cluttered scenes using simulated robot experiments. Our framework only collects labeled instance grasping data from simulation, and uses a domain-adversarial loss to transfer the trained model to real robots using indiscriminate grasping data from the simulation and the real world. We also presented a deep neural network, which takes monocular RGB images and instance segmentation mask as input, and predicts the probability of successfully grasping the specified object for each candidate motor command. We demonstrated our method on a real robot platform for grasping household dishware objects that were never seen during the training in simulation. And we showed that our model outperforms alternative model architectures and indiscriminate grasping baseline. 

One limitation of using Mask R-CNN to predict segmentation masks is that it does not work well when objects are occluded by the robot gripper, which happens frequently during grasping. In addition, we cannot run Mask R-CNN for every single time step due to its computational expenses. Therefore, we only predict the segmentation mask for the initial RGB image. While we find our model can extracts effective information from the single segmentation mask, it would be interesting to see how the instance grasping performance can be further improved by updating the segmentation masks for the later time steps.

In our work, we use the segmentation mask as the input modality for specifying the target object. There can be other options for specifying the target objects such as detection bounding boxes and user-specified pixels on the object. It will be straightforward to extend our method to these alternative input modalities. Although we choose instance grasping and indiscriminate grasping in our problem setup, our framework can potentially be applied to other tasks that share similar properties, \eg robot picking and placing. Incorporating our framework to other robot tasks will be an exciting avenue for future works.

%%%%%%%%%%%%%%%%%%%%%%%%%%%%%%%%%%%%%%%%%%%%%%%%%%%%%%%%%%%%%%%%%%%%%%%%%%%%%%%%
\section*{ACKNOWLEDGMENT}
We thank Adrian Li, Peter Pastor, Ian Wilkes, Kurt Konolige for contributions to the software and hardware development for the grasping infrastructures. We thank Alex Irpan, Paul Wohlhart, Konstantinos Bousmalis for discussions on training and implementation of the domain classifier. We thank John-Michael Burke for the supports of the real-world robot experiments.

\addtolength{\textheight}{-12cm}   % This command serves to balance the column lengths
                                  % on the last page of the document manually. It shortens
                                  % the textheight of the last page by a suitable amount.
                                  % This command does not take effect until the next page
                                  % so it should come on the page before the last. Make
                                  % sure that you do not shorten the textheight too much.

% {\small
% \bibliographystyle{bibtex/IEEEtran}
% \bibliography{bibtex/instance_grasping}
% }

\end{document}